\documentclass[11pt,a4paper]{article}
\usepackage{acl}
\usepackage{times}
\usepackage{latexsym}
\usepackage{amsthm}

\usepackage{amsmath}
\usepackage{graphicx}
\usepackage{algorithm}
\usepackage[noend]{algpseudocode}

\usepackage{multirow}
\usepackage{adjustbox}
\usepackage{bbm}
\usepackage{wrapfig,lipsum}
\usepackage{xspace}
\usepackage{booktabs}       
\usepackage[normalem]{ulem}
\usepackage{makecell}
\usepackage{amssymb}
\usepackage{pifont}
\newcommand{\cmark}{\ding{51}}%
\newcommand{\xmark}{\ding{55}}%

\usepackage{microtype}

\usepackage{cleveref}
\crefformat{section}{\S#2#1#3}
\crefformat{subsection}{\S#2#1#3}
\crefformat{subsubsection}{\S#2#1#3}
\crefrangeformat{section}{\S\S#3#1#4 to~#5#2#6}
\crefmultiformat{section}{\S\S#2#1#3}{ and~#2#1#3}{, #2#1#3}{ and~#2#1#3}
\usepackage{refstyle}
\Crefformat{figure}{#2Fig.~#1#3}
\Crefmultiformat{figure}{Figs.~#2#1#3}{ and~#2#1#3}{, #2#1#3}{ and~#2#1#3}
\Crefformat{table}{#2Tab.~#1#3}
\Crefmultiformat{table}{Tabs.~#2#1#3}{ and~#2#1#3}{, #2#1#3}{ and~#2#1#3}
\Crefformat{appendix}{Appx.~\S#2#1#3}
\crefformat{algorithm}{Alg.~#2#1#3}
\crefformat{equation}{Eq.~#2#1#3}
\crefformat{definition}{Def.~#2#1#3}

\newcommand{\stitle}[1]{\vspace{1ex} \noindent{\bf #1.}}

\newcommand{\modelname}{\textsc{CoRE}\xspace}

\title{Should We Rely on Entity Mentions for Relation Extraction?\\
Debiasing Relation Extraction with Counterfactual Analysis
}

\author{
Yiwei Wang$^1$ \ \ \ \ Muhao Chen$^2$ \ \ \ \ Wenxuan Zhou $^2$ \ \ \ \ Yujun Cai$^3$ \ \ \ \  Yuxuan Liang$^1$ \ \ \ \ \\ \textbf{Dayiheng Liu$^4$ \ \ \ \  Baosong Yang$^4$ \ \ \ \ Juncheng Liu$^1$ \ \ \ \ Bryan Hooi$^1$} \\ 
$^1$ National University of Singapore \quad
$^2$ University of Southern California  \\
$^3$ Nanyang Technological University \quad 
$^4$Alibaba Group \\
\texttt{wangyw\_seu@foxmail.com}
}

\date{}

\begin{document}
\maketitle
\begin{abstract}
Recent literature focuses on utilizing the entity information in the sentence-level relation extraction (RE), but this risks leaking superficial and spurious clues of relations.
As a result, RE still suffers from unintended \textbf{entity bias}, i.e., the spurious correlation between \textbf{entity mentions (names)} and relations.
Entity bias can mislead the RE models to extract the relations that do not exist in the text.
To combat this issue, some previous work masks the entity mentions to prevent the RE models from over-fitting entity mentions.
However, this strategy degrades the RE performance because it loses the semantic information of entities.
In this paper, we propose the \modelname (\textbf{Counterfactual Analysis based Relation Extraction}) debiasing method that guides the RE models to focus on the main effects of \textbf{textual context} without losing the entity information.
We first construct a causal graph for RE, which models the dependencies between variables in RE models.
Then, we propose to conduct counterfactual analysis on our causal graph to distill and mitigate the entity bias, that captures the causal effects of specific entity mentions in each instance.
Note that our \modelname method is model-agnostic to debias existing RE systems during inference without changing their training processes.
Extensive experimental results demonstrate that our \modelname yields significant gains on both effectiveness and generalization for RE.
The source code is provided at: \url{https://github.com/vanoracai/CoRE}.
\end{abstract}

\section{Introduction}\label{sec:int}

Sentence-level relation extraction (RE) is an important step to obtain a structural perception of unstructured text \cite{distiawan2019neural} by extracting relations between \textbf{entity mentions (names)} from the \textbf{textual context}.
From human oracle, textual context should be the main source of information that determines the ground-truth relations between entities.
Consider a sentence ``\textit{\uwave{Mary} gave birth to \underline{Jerry}.}''\footnote{We use \underline{underline} and \uwave{wavy line} to denote subject and object respectively by default.}. 
Even if we change the entity mentions from \textit{`Jerry'} and \textit{`Mary'} to other people's names, the relation `parents' still holds between the subject and object as described by the textual context \textit{``gave birth to''}.

\begin{figure*}[!tb]
	\centering
	\includegraphics[width=1\linewidth]{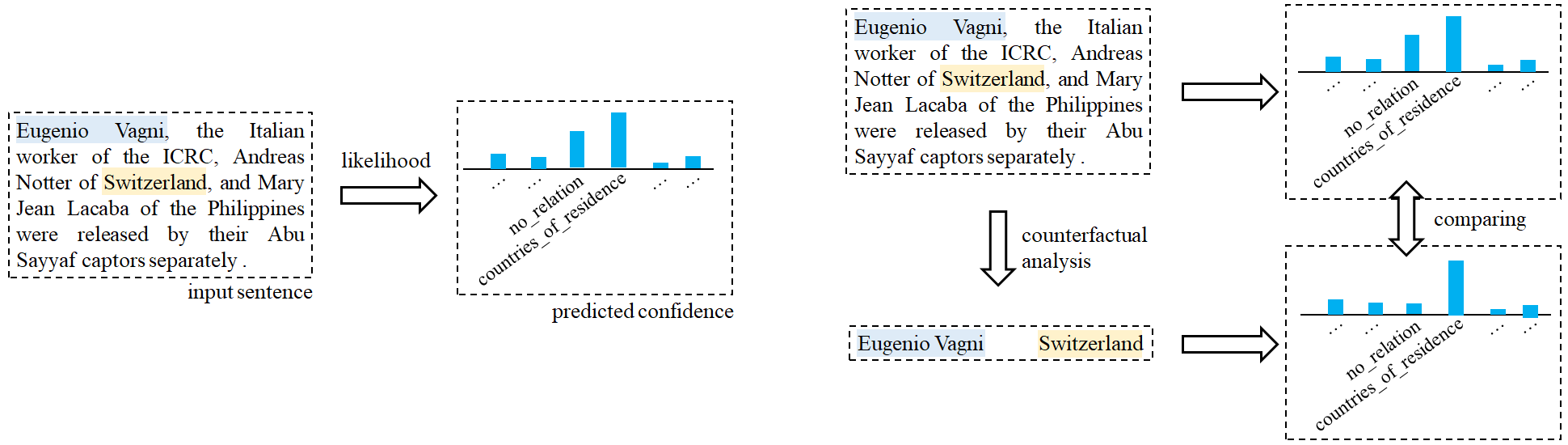}
	\caption{
		(\textit{left}) An example of RE produced by LUKE \cite{yamada-etal-2020-luke}.
		In the input sentence, the subject is in blue and the object is in yellow.
		The ground-truth relation between the subject and object is ``\textit{no\_relation}'', since there is not any relation reflected by the textual context. 
		(\textit{right}) Our proposed counterfactual analysis for RE, which compares the original prediction (upper) with the counterfactual one (lower) to mitigate the entity bias.
		\label{fig:1}}
\end{figure*}

Recently, some work aims to utilize entity mentions for RE \cite{yamada-etal-2020-luke,zhou2021improved}, which, however, leak superficial and spurious clues about the relations \cite{zhang2018graph}.
In our work, we observe that entity information can lead to \textbf{biased} relation prediction by misleading RE models to extract relations that do not exist in the text.
\Cref{fig:1} visualizes a relation prediction from a state-of-the-art RE model \cite{alt2020tacred} (see more examples in \Cref{tab:case}).
Although the context describes no relation between the highlighted entity pair, the model extracts the relation as ``\textit{countries\_of\_residence}''.
Such an erroneous result can come from the spurious correlation between entity mentions and relations, or the \textbf{entity bias} in short.
For example, if the model sees the relation ``\textit{countries\_of\_residence}'' many more times than other relations when the object entity is \textit{Switzerland} during training, the model can associate this relation with \textit{Switzerland} during inference even though the relation does not exist in the text.

To combat this issue, some work \cite{zhang2017position,zhang2018graph} proposes masking entities to prevent the RE models from over-fitting entity mentions.
On the other hand, some other work \cite{peng2020learning,zhou2021improved} finds that this strategy degrades the performance of RE because it loses the semantic information of entities.

For both machines and humans, RE requires a combined understanding of textual context and entity mentions \cite{peng2020learning}.
Humans can avoid the entity bias and make unbiased decisions by correctly referring to the textual context that describes the relation.
The underlying mechanism is \textit{causality-based} \cite{van2015cognitive}: humans identify the relations by pursuing the main causal effect of the textual context instead of the side-effect of entity mentions.
In contrast, RE models are usually \textit{likelihood-based}: the prediction is analogous to looking up the entity mentions and textual context in a huge likelihood table, interpolated by training \cite{tang2020unbiased}.
In this paper, our idea is to teach RE models to distinguish between the effects from the textual context and entity mentions through counterfactual analysis \cite{pearl2018causal}:

\noindent\textbf{Counterfactual analysis:} \textit{If I had not seen the textual context, would I still extract the same relation?}

\noindent The counterfactual analysis essentially gifts humans the hypothesizing abilities 
to make decisions 
collectively based on the textual context and entity mentions, as well as to introspect whether the decision is deceived (see \Cref{fig:1}).
Specifically, we are essentially comparing the original instance with a counterfactual instance, where only the textual context is wiped out, while keeping the entity mentions untouched.
By doing so, we can focus on the main effects of the textual context without losing the entity information.

In our work, we propose a novel model-agnostic paradigm for debiasing RE, namely \modelname (\underline{Co}unterfactual analysis based \underline{R}elation \underline{E}xtraction), which adopts the counterfactual analysis to mitigate the spurious influence of the entity mentions.
Specifically, \modelname does not touch the training of RE models, i.e., it allows a model to be exposed to biases on the original training set.
Then, we construct a causal graph for RE to analyze the dependencies between variables in RE models, which acts as a ``roadmap'' for capturing the causal effects of textual context and entity mentions.
To rectify the test instances from the 
potentially biased prediction, in inference, \modelname ``imagines'' the counterfactual counterparts on our causal graph to distill the biases.
Last but not least, \modelname performs a bias mitigation operation with adaptive weights to produce a debiased decision for RE.

We highlight that \modelname is a flexible debiasing method that is applicable to popular RE models without changing their training processes.
To evaluate the effectiveness of \modelname, we perform extensive experiments on public benchmark datasets. 
The results demonstrate that our proposed method can significantly improve the effectiveness and generalization of the popular RE models by mitigating the biases in an entity-aware manner.

\section{Related Work}
\stitle{Sentence-level relation extraction}
Early research efforts \cite{nguyen-grishman-2015-relation,wang-etal-2016-relation,zhang2017position} train RE models from scratch based on lexicon-level features.
The recent RE work fine-tunes pretrained language models (PLMs; \citealt{devlin-etal-2019-bert,liu2019roberta}).
For example, K-Adapter~\cite{wang2020k} fixes the parameters of the PLM and uses feature adapters to infuse factual and linguistic knowledge.
Recent work focuses on utilizing the entity information for RE \cite{zhou2021improved,yamada-etal-2020-luke}, but this leaks superficial and spurious clues about the relations \cite{zhang2018graph}.
Despite the biases in existing RE models, scarce work has discussed the spurious correlation between entity mentions and relations that causes such biases.
Our work investigates this issue and proposes \modelname to debias RE models for higher effectiveness.

\stitle{Debiasing for Natural Language Processing}
Debiasing is a fundamental problem in machine learning \cite{torralba2011unbiased}.
For natural language processing (NLP), some work performs data re-sampling to prevent models from capturing the unintended bias in training \cite{dixon2018measuring,geng2007boosting,kang2016noise,rayhan2017cusboost,nguyen2011borderline}.
Alternatively, \citet{wei2019eda} and \citet{qian2020enhancing} develop data augmentation for debiasing.
Some recent work debiases the NLP models based on causal inference \cite{qian2021counterfactual,nan2021uncovering}.
In RE, how to deal with the entity bias is also an important problem.
For example, PA-LSTM \cite{zhang2017position} masks the entity mentions with special tokens to prevent RE models from over-fitting entity names, which was also adopted by C-GCN~\cite{zhang2018graph} and SpanBERT \cite{joshi-etal-2020-spanbert}.
However, masking entities loses the semantic information of entities and leads to performance degradation.
Different from it, our \modelname model tackles entity biases based on structured causal models.
In this way, we debias the RE models to focus on the textual context without losing the entity information.

\begin{figure}[!tb]
	\centering
	\includegraphics[width=0.7\linewidth] {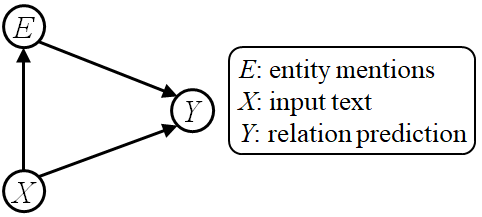}
	\caption{
		The causual graph of RE models.
		\label{fig:3}}
\end{figure}

\section{Methodology}
Sentence-level relation extraction (RE) aims to extract the relation between a pair of entities mentioned from a sentence.
We propose \modelname (counterfactual analysis based Relation Extraction) as a model-agnostic technique to endow existing RE models with unbiased decisions during inference.
\modelname follows the regular training process of existing work regardless of the bias from the entity mentions.
During inference, \modelname post-adjusts the biased prediction according to the effects of the bias.
\modelname can be flexibly incorporated into popular RE models to improve their effectiveness and generalization based on the counterfactual analysis without re-training the model.

In this section, we first formulate the existing RE models in the form of a causal graph.
Then, we introduce our proposed bias distillation method to distill the entity bias with our designed counterfactual analysis.
We conduct an empirical analysis to analyze how heavily the existing RE models rely on the entity mentions to make decisions. 
Finally, we mitigate the distilled bias from the predictions of RE models to improve their effectiveness.

\subsection{Causality of Relation Extraction}\label{sec:3_1}
In order to perform causal intervention, we first formulate the \textbf{causal graph} \cite{pearl2016causal,pearl2018book}, a.k.a., structural causal model, for the RE models as \Cref{fig:3}, which sheds light on how the textual context and entity mentions affect the RE predictions.
The causal graph is a directed acyclic graph $\mathcal{G} = \{\mathcal{V}, \mathcal{E}\}$, indicating how a set of variables $\mathcal{V}$ interact with each other through the causal relations behind the data and how variables obtain their values, e.g., $(E, X) \rightarrow Y$ in \Cref{fig:3}.
Before we conduct counterfactual analysis that deliberately manipulates the values of nodes and prunes the causal graph, we first revisit the conventional RE systems in the graphical view.

The causal graph in \Cref{fig:3} is applicable to a variety of RE models 
and imposes no constraints on the detailed implementations.
Node $X$ is the input text.
On the edge $X \rightarrow E$, we obtain the spans of subject and object entities as node $E$ through NER or human annotations \cite{zhang2017position}.
For example, in the aforementioned sentence $X =$ ``\textit{\uwave{Mary} gave birth to \underline{Jerry}.}'', the entities are $E=$ [\textit{'Mary', 'Jerry'}].

On the edges $(X, E) \rightarrow Y$, existing RE models take different designs.
For example, C-GCN~\cite{zhang2018graph} obtains the relation prediction $Y$ by encoding entity mentions $E$ on the pruned dependency tree of $X$ using a graph convolutional network.
IRE \cite{zhou2021improved} uses PLMs as the encoder for $X$, and marks the entity information of $E$ with special tokens to utilize the entity information.

\subsection{Bias Distillation}\label{sec:3_2}
Based on our causal graph in \Cref{fig:3}, we diagnose how the entity bias affects inference.
After training, the causal dependencies among the variables are learned in terms of the model parameters.
The entity bias can mislead the models to make wrong predictions while ignoring the actual relation-describing textual context in $X$, i.e., biased towards the causal dependency: $E \rightarrow Y$.

\begin{figure}[!tb]
	\centering
	\includegraphics[width=1.0\linewidth]{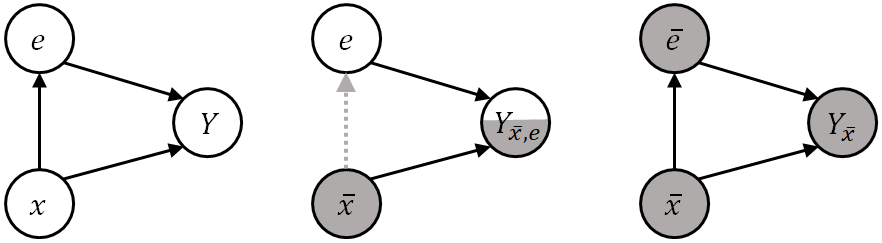}
	\caption{
		The original causal graph of RE models (left) together with its two counterfactual alternates for the entity bias (middle) and label bias (right).
		The shading indicates the mask of corresponding variables.
		\label{fig:5}}
\end{figure}

The conventional biased prediction can only see the output $Y$ of the entire graph given a sentence $X$, ignoring how specific entity mentions affect the relation prediction.
However, causal inference encourages us to think out of the black box.
From the graphical point of view, we are no longer required to execute the entire causal graph as a whole.
In contrast, we can directly manipulate the nodes and observe the output.
The above operation is termed \textbf{intervention} in causal inference, which we denote as $do(\cdot)$. 
It wipes out all the incoming links of a node and demands it to take a certain value.

We distill the entity bias by intervention and its induced \textbf{counterfactual}.
The counterfactual means ``counter to the facts'', and takes one step that further assigns the hypothetical combination of values to variables.
For example, we can remove the input textual context by masking $X$, but maintain $E$ as the original entity mentions, as if $X$ still exists.

We will use the input text $X$ as our control variable where the intervention is conducted, aiming to assess its effects, due to the fact that there would not be any valid relation between entities in $E$ if the input text $X$ is empty.
We denote the output logits $Y$ after the intervention $X = \bar{x}$ as follows:
\begin{equation}
Y_{\bar{x}} = Y(do(X = \bar{x})).
\end{equation}
Following the above notation, the original prediction $Y$, i.e., can be re-written as $Y_x$.

To distill the entity bias, we conduct the intervention $do(X = \bar{x})$ on $X$, while keeping the variable $E$ as the original $e$, as if the original input text $x$ had existed.
Specifically, we mask the tokens in $x$ to produce $\bar{x}$ but keep the entity mentions $e$ as original, so that the textual context is removed and the entity information is maintained.
Accordingly, the counterfactual prediction is denoted as $Y_{\bar{x}, e}$ (see \Cref{fig:5}).
In this case, since the model cannot see any textual context in the factual input $x$ after the intervention $\bar{x}$, but  still has access to the original entity mentions $e$ as the inputs, the prediction $Y_{\bar{x}, e}$ purely reflects the influence from $e$.
In other words, $Y_{\bar{x}, e}$ refers to the output, i.e., a probability distribution or a logit vector, where only the entity mentions are given as the input without textual context.

\begin{figure}[!tb]
	\centering
	\includegraphics[width=0.95\linewidth]{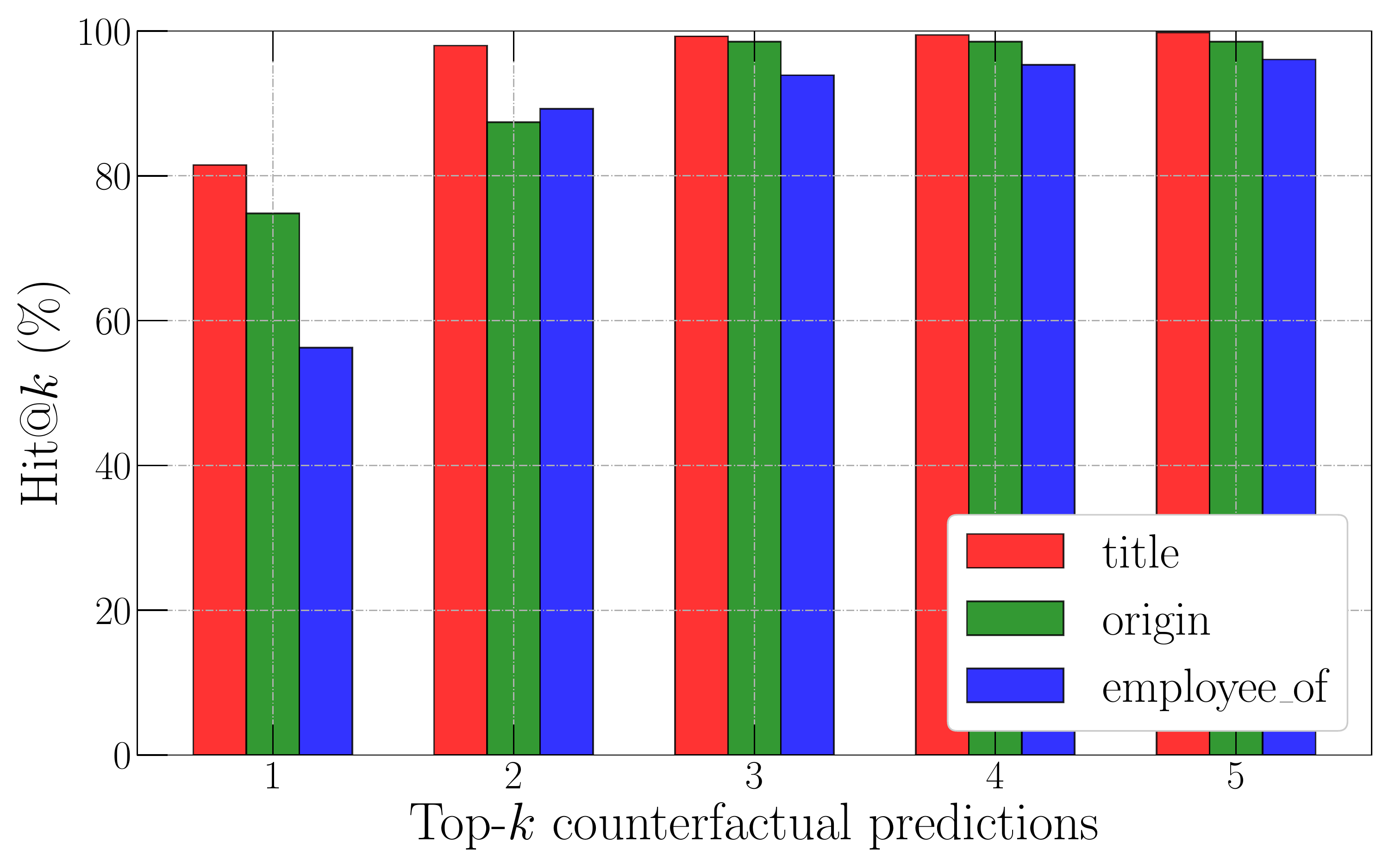}
	\caption{
		Hit$@k$ (y-axis) is the fraction of the test instances, that have the original relation prediction $\arg\max_c Y_x[c]$ ranked in the top $k$ most confident relations of the counterfactual prediction $Y_{\bar{x}, e}$.
		We report Hit$@k$ of the model IRE$_{\mathrm{RoBERTa}}$ on the test instances when the original relation prediction is \textit{title}, \textit{employee\_of}, or \textit{origin}.
		\label{fig:4}}
\end{figure}

To investigate how heavily the state-of-the-art models rely on the entity mentions for RE, we conduct an empirical study to compare the original prediction $Y_{x}$ and the counterfactual one $Y_{\bar{x}, e}$.
Specifically, we calculate the fraction of the test instances (y-axis) that have the original relation prediction $\arg\max_c Y_x[c]$ ranked in the top $k$ most confident relations of the counterfactual prediction $Y_{\bar{x}, e}$.
This fraction is termed as Hit$@ k$.

We present Hit$@ k$ for IRE$_{\mathrm{RoBERTa}}$ \cite{zhou2021improved}, a state-of-the-art RE model, in \Cref{fig:4} on the test instances when the original relation prediction is \textit{title}, \textit{employee\_of}, or \textit{origin}.
Higher Hit$@ 1$ means that for more instances, the model infers the same relation given only the entity mentions no matter whether the textual context is given, which imply stronger causal effects from the entity mentions $Y_{\bar{x}, e}$, i.e., the models rely more heavily on the entity mentions for RE.

We observe that when $k = 1$, the Hit$@ 1$ is more than 50\%, which implies that the model typically extracts the same relations even without textual context on more than a half of the instances.
For a larger $k$, the Hit$@ k$ increases significantly and reaches more than 80\% for $k \ge 2$.
These observations imply a promising but embarrassing result: the state-of-the-art model relies on the entity bias for RE on many instances.
The entity bias reflected by $Y_{\bar{x}, e}$ can lead to the wrong extraction if the relation implied by the entity mentions does not exist in the input text.
This poses a challenge to the generalization of RE models, as validated by our experimental results (\Cref{sec:4-3}).

In addition to $Y_{\bar{x}, e}$ that reflects the causal effects of entity mentions, there is another kind of bias not conditioned on the entity mentions $e$, but reflecting the general bias in the whole dataset, which is $Y_{\bar{x}}$.
$Y_{\bar{x}}$ corresponds to the counterfactual inputs where both textual context and entity mentions are removed.
In this case, since the model cannot access any information from the input after this removal, $Y_{\bar{x}}$ naturally reflects the label bias that exists in the model from the biased training.
The causal graphic views of the original prediction $Y_{x}$, the counterfactual $Y_{\bar{x}, e}$ for the entity bias, and $Y_{\bar{x}}$ for the label bias are visualized in \Cref{fig:5}.

\subsection{Bias Mitigation}\label{sec:3_2}
As we have discussed in \Cref{sec:int}, instead of the static likelihood that tends to be biased, the unbiased relation prediction lies in the difference between the observed outcome $Y_{x}$ and its counterfactual predictions $Y_{\bar{x}, e},Y_{\bar{x}}$.
The latter two are the biases that we want to mitigate from the relation prediction.

Intuitively, the unbiased prediction that we seek is the linguistic stimuli from blank to the observed textual context with specific relation descriptions, but not merely from the entity bias.
The context-specific clues of the relations are key to the informative unbiased predictions, because even if the overall prediction is biased towards the relation ``\textit{schools\_attended}'' due to the object entity like ``\textit{Duke University}'', the textual context \textit{``work at''} indicates the relation as \textit{``employee\_of''} rather than \textit{``schools\_attended''}.

Our final goal is to use the direct effect of the textual context from $X$ to $Y$ for debiased prediction, mitigating (denoted as \textbackslash) the label bias and the entity bias from the prediction: $Y_x$  \textbackslash $Y_{\bar{x}, e}$ \textbackslash $Y_{\bar{x}}$, so as to block the spread of the biases from training to inference.
The debiased prediction via bias mitigation can be formulated via the conceptually simple but empirically effective element-wise subtraction operation:
\begin{equation}\label{eq:debias}
    Y_{\mathrm{final}} = Y_x - \lambda_1 Y_{\bar{x}, e} - \lambda_2 Y_{\bar{x}},
\end{equation}
where $\lambda_1$ and $\lambda_2$ are two independent hyper-parameters balancing the terms for mitigating entity and label biases respectively.
Note that the bias mitigation in \Cref{eq:debias} for the entity and label biases correspond to Total Direct Effect (TDE) and Total Effect (TE) in causal inference \cite{tang2020unbiased,vanderweele2015explanation,pearl2009causality} respectively.
We adaptively set the values of $\lambda_1$ and $\lambda_2$ for different datasets based on the grid beam search \cite{hokamp2017lexically} in a scoped two dimensional space:
\begin{equation}\label{eq:search}
    \lambda_1^{\star}, \lambda_2^{\star} = \arg\max_{\lambda_1, \lambda_2}\psi(\lambda_1, \lambda_2) \ \ \lambda_1, \lambda_2 \in [a, b],
\end{equation}
where $\psi$ is a metric function (e.g., F1 scores) for evaluation, $a, b$ are the boundaries of the search range.
We search the values of $\lambda_1, \lambda_2$ once on the validation set, and use the fixed values for inference on all testing instances.
Since the entity types can restrict the candidate relations \cite{lyu2021relation}, we use the entity type information, if available, to restrict the candidate relations for inference, which strengthens the effects of entity types for relation extraction.

Overall, the proposed \modelname replaces the conventional one-time prediction with $Y_{\mathrm{final}}$ to produce the debiased relation predictions, which essentially ``thinks'' twice: one for the original observation $Y_x$, the other for hypothesized $Y_{\bar{x}}, Y_{\bar{x}, e}$.

\begin{table}[tb!]
	\centering
	\begin{adjustbox}{width=\linewidth}
		
		\begin{tabular}{@{}l| c c c c @{}}
			\toprule
			\textbf{Dataset}
			& $\#$\textbf{Train}	
			& $\#$\textbf{Dev}
			& $\#$\textbf{Test}
			& $\#$\textbf{Classes} \\ 
			\midrule
			\midrule
			TACRED & 68,124 & 22,631 & 15509 & 42 \\
			SemEval & 6,507 & 1,493 & 2,717 & 19\\
			Re-TACRED & 58,465 & 19,584 & 13418 & 40 \\
			TACRED-Revisit & 68,124 & 22,631 & 15509 & 42 \\
			\bottomrule
		\end{tabular}
		
	\end{adjustbox}
		\caption{Statistics of datasets.}	\label{tab:data}

	\vspace{-0.4cm}
\end{table}

\section{Experiments}
In this section, we 
evaluate the performance of our \modelname methods 
when applied to RE models.
We compare our methods against a variety of strong baselines on the task of sentence-level RE.
Our experimental settings closely follow those of the previous work \cite{zhang2017position,zhou2021improved,nan2021uncovering} to ensure a fair comparison.

\subsection{Experimental Settings}

\stitle{Datasets}
We use four widely-used RE benchmarks: TACRED \cite{zhang2017position}, SemEval \cite{hendrickx2019semeval}, TACRED-Revisit \cite{alt2020tacred}, and Re-TACRED \cite{stoica2021re} for evaluation.
TACRED contains over 106k mention pairs drawn from the yearly TAC KBP challenge.
\cite{alt2020tacred} relabeled the development and test sets of TACRED.
Re-TACRED is a 
further relabeled version of TACRED after refining its label definitions.
The statistics of these datasets are shown in \Cref{tab:data}.

\begin{table*}[tb!]
    \small
	\centering
		\begin{tabular}{@{}l c c c c c c c c@{}}
			\toprule
			\textbf{Method}
			& \textbf{TACRED}
			& \textbf{TACRED-Revisit}
			& \textbf{Re-TACRED}
			& \textbf{SemEval}\\ 
			\midrule
			\midrule
			C-SGC \cite{wu2019simplifying} & 52.1 & 62.8 & 69.8 & 71.3 \\
			SpanBERT \cite{joshi-etal-2020-spanbert} & 55.7 & 65.1 & 74.1 & 74.9 \\
			CP \cite{peng2020learning} & 56.8 & 67.1 & 78.1 & 79.6 \\
			RECENT \cite{lyu2021relation} & 63.3 & 70.5 & 81.1 & 74.6 \\
			KnowPrompt \cite{chen2021knowprompt} & 57.6 & 68.7 & 79.0 & 81.8 \\
			IRE$_{\mathrm{BERT}}$ \cite{zhou2021improved} & 59.2 & 68.4 & 78.6 & 79.1 \\
			\midrule
			\midrule
			LUKE \cite{yamada-etal-2020-luke} & 58.8 & 67.5 & 80.2 & 82.1 \\
			\midrule
			LUKE + Resample \cite{burnaev2015influence} & 59.3 & 68.2 & 80.5 & 82.5 \\
			LUKE + Focal \cite{lin2017focal} & 59.1 & 67.7 & 80.3 & 82.4 \\
			LUKE + CFIE \cite{nan2021uncovering} & 59.8 & 68.0 & 80.4 & 82.2 \\
			LUKE + Entity Mask \cite{zhang2017position}  & 57.9 & 67.0 & 79.5 & 82.0 \\
			LUKE + \modelname & 61.7 & 70.2 & 81.6 & \textbf{83.6} \\
			\midrule
			\midrule
			IRE$_{\mathrm{RoBERTa}}$ \cite{zhou2021improved} & 63.1 & 70.6 & 81.5 & 81.4  \\
			\midrule
			IRE$_{\mathrm{RoBERTa}}$ + Resample \cite{burnaev2015influence} & 63.3 & 71.0 & 81.9 & 81.6 \\
			IRE$_{\mathrm{RoBERTa}}$ + Focal \cite{lin2017focal} & 62.9 & 70.7 & 81.2 & 81.1 \\
			IRE$_{\mathrm{RoBERTa}}$ + CFIE \cite{nan2021uncovering} & 63.3 & 70.9 & 81.6 & 81.7 \\
			IRE$_{\mathrm{RoBERTa}}$ + Entity Mask \cite{zhang2017position} & 61.4 & 69.3 & 79.6 & 81.2 \\
			IRE$_{\mathrm{RoBERTa}}$ + \modelname & \textbf{64.4} & \textbf{71.8} & \textbf{82.8} & 82.3 \\
			\bottomrule
		\end{tabular}

	\caption{F1-macro scores (\%) of RE on the test sets of TACRED, TACRED-Revisit, Re-TACRED, and SemEval. The best results in each column are highlighted in \textbf{bold} font.}	\label{tab:1}
	\vspace{-0.4cm}
\end{table*}

We use the widely-used F1-macro score as the main evaluation metric \cite{nan2021uncovering}, which is the balanced harmonic mean of precision and recall, as well as F1-micro for a more comprehensive evaluation.
F1-macro is more suitable than F1-micro to reflect the extent of biases, especially for the highly-skewed cases, since F1-macro is evenly influenced by the performance in each category, i.e. category-sensitive, but F1-micro simply gives equal weights to all instances \cite{kim2019small}.

\begin{table*}[tb!]
    \small
    
	\centering
	
		\begin{tabular}{@{}l c c c c c c c c@{}}
			\toprule
			\textbf{Method}
			& \textbf{TACRED}
			& \textbf{TACRED-Revisit}
			& \textbf{Re-TACRED}
			& \textbf{SemEval}\\ 
			\midrule
			\midrule
			LUKE \cite{yamada-etal-2020-luke} & 72.7 & 80.6 & 90.3 & 87.8 \\
			\midrule
			LUKE + Resample  \cite{burnaev2015influence} & 73.1 & 80.9 & 90.5 & 87.9 \\
			LUKE + Focal \cite{lin2017focal} & 72.9 & 80.7 & 90.4 & 87.6 \\
			LUKE + CFIE \cite{nan2021uncovering} & 73.3 & 80.8 & 90.5 & 88.0\\
			LUKE + Entity Mask \cite{zhang2017position} & 72.3 & 80.4 & 90.1 & 87.5 \\
			LUKE + \modelname & \textbf{74.6} & \textbf{81.4} & \textbf{90.9} & \textbf{88.7} \\
			\bottomrule
		\end{tabular}

	\caption{F1-micro scores (\%) of RE on the test sets of TACRED, TACRED-Revisit, Re-TACRED, and SemEval. The best results in each column are highlighted in \textbf{bold} font.}\label{tab:1}
	\vspace{-0.4cm}
\end{table*}

\begin{table}[tb!]
	\centering

	\begin{adjustbox}{width=\linewidth}
		\begin{tabular}{@{}l c c @{}}
			\toprule
			\textbf{Method}
			& \textbf{TACRED}
			& \textbf{Re-TACRED}\\ 
			\midrule
			\midrule
			LUKE \cite{yamada-etal-2020-luke} & 51.9 & 65.3  \\
			\midrule
			w/ Resample \cite{burnaev2015influence} & 53.2 & 66.7 \\
			w/ Focal \cite{lin2017focal} & 52.4 & 65.9 \\
			w/ CFIE \cite{nan2021uncovering} & 52.1 & 65.6 \\
			w/ Entity Mask \cite{zhang2017position} &  54.5 & 67.1\\
			w/ \modelname (ours) & \textbf{69.3} & \textbf{83.1} \\
			\midrule
			\midrule
			IRE$_{\mathrm{RoBERTa}}$ \cite{zhou2021improved} & 56.4 & 68.1 \\
			\midrule
			w/ Resample \cite{burnaev2015influence} & 58.1 & 70.3 \\
			w/ Focal \cite{lin2017focal} & 56.8 & 68.7 \\
			w/ CFIE \cite{nan2021uncovering} & 57.1 & 68.4 \\
			w/ Entity Mask \cite{zhang2017position} & 57.3 & 68.9 \\
			w/ \modelname (ours) & \textbf{73.6} & \textbf{85.4} \\
			\bottomrule
		\end{tabular}
	\end{adjustbox}

	\caption{F1-macro scores (\%) of RE on the challenging test sets of TACRED and Re-TACRED, in which the relations implied by the entity mentions do not exist in the textual context.
	`w\/' denotes `with'.
	The best results in each column are highlighted in \textbf{bold} font.}\label{tab:chall}
	\vspace{-0.4cm}
\end{table}

\begin{table*}[tb!]
    \small
	\centering
		\begin{tabular}{@{}l c c c c c c c c@{}}
			\toprule
			\textbf{Method}
			& \textbf{TACRED}
			& \textbf{TACRED-Revisit}
			& \textbf{Re-TACRED}
			& \textbf{SemEval}\\ 
			\midrule
			\midrule
			IRE$_{\mathrm{RoBERTa}}$ \cite{zhou2021improved} & 61.2 & 59.3 & 57.5 & 54.1 \\
			\midrule
			IRE$_{\mathrm{RoBERTa}}$ + Resample \cite{burnaev2015influence} & 60.5 & 58.4 & 56.8 & 53.5 \\
			IRE$_{\mathrm{RoBERTa}}$ + Focal \cite{lin2017focal} & 60.9 & 58.9 & 57.1 & 53.7 \\
			IRE$_{\mathrm{RoBERTa}}$ + CFIE \cite{nan2021uncovering} & 60.1 & 57.8 & 56.2 & 52.9\\
			IRE$_{\mathrm{RoBERTa}}$ + Entity Mask \cite{zhang2017position} & 61.5 & 60.1 & 57.3 & 54.2 \\
			IRE$_{\mathrm{RoBERTa}}$ + \modelname & \textbf{57.3} & \textbf{55.6} & \textbf{54.3} & \textbf{50.8} \\
			\bottomrule
		\end{tabular}

	\caption{Experimental results (unfairness; \%) of Relation Extraction on the test sets of TACRED, TACRED-Revisit, Re-TACRED, and SemEval (lower is better).
	The best results in each column are highlighted in \textbf{bold} font. \label{tab:fair}
	}
\end{table*}

\stitle{Compared methods}
We take the following RE models into comparison.
(1) \textbf{C-SGC}~\cite{wu2019simplifying} simplifies GCN, and combines it with LSTM, leading to improved performance over each method alone. 
(2) \textbf{SpanBERT}~\cite{joshi-etal-2020-spanbert} extends BERT by introducing a new pretraining objective of continuous span prediction.
(3) \textbf{CP}~\cite{peng2020learning} is an entity-masked contrastive pre-training framework for RE.
(4) \textbf{RECENT} \cite{lyu2021relation} restricts the candidate relations based on the entity types.
(5) \textbf{KnowPrompt} \cite{chen2021knowprompt} is Knowledge-aware Prompt-tuning approach.
(6) \textbf{LUKE}~\cite{yamada-etal-2020-luke} pretrains the language model on both large text corpora and knowledge graphs and further proposes an entity-aware self-attention mechanism. 
(7) \textbf{IRE}~\cite{zhou2021improved} proposes an improved entity representation technique in the data preprocessing.

Among the above RE models, we apply our \modelname on LUKE and IRE.
To demonstrate the effectiveness of debiased inference, we also compare with the following debiasing techniques that are applied to the same two RE models.
(1) \textbf{Focal} \cite{lin2017focal} adaptively reweights the losses of different instances so as to focus on the hard ones. 
(2) \textbf{Resample} \cite{burnaev2015influence} up-samples rare categories by the inversed sample fraction during training.
(3) \textbf{Entity Mask} \cite{zhang2017position}: masks the entity mentions with special tokens to reduce the over-fitting on entities.
(4) \textbf{CFIE} \cite{nan2021uncovering} is also a causal inference method.
In contrast to our method, CFIE strengthens the causal effects of entities by masking entity-centric information in the counterfactual predictions.

\stitle{Model configuration}
For the hyper-parameters of the considered baseline methods, e.g., the batch size, the number of hidden units, the optimizer, and the learning rate, we set them as suggested by their authors.
For the hyper-parameters of our \modelname method, we set the search range of the hypermeters in \Cref{eq:search} as $[-2, 2]$ and the search step $0.1$.
For all experiments, we report the median $F1$ scores of five runs of training using different random seeds.

\subsection{Overall Performance}\label{sec:4-1}
We implement our \modelname with LUKE and IRE$_{\mathrm{RoBERTa}}$.
\Cref{tab:1} reports the RE results on the TACRED, TACRED-Revisit, Re-TACRED, and SemEval datasets.
Our \modelname method improves the F1-macro scores of LUKE by 4.9\% on TACRED, 4.0\% on TACRED-Revisit, 1.7\% on Re-TACRED, and 1.7 on SemEval, and improves IRE$_{\mathrm{RoBERTa}}$ by 1.2\% on TACRED, 1.4\% on TACRED-Revisit, 0.9\% on Re-TACRED, and 1.8\% on SemEval.
As a result, our \modelname achieves substantial improvements for LUKE and IRE$_{\mathrm{RoBERTa}}$, and enables them to outperform the baseline methods.
Additionally, we report the experimental results in terms of F1-micro scores in \Cref{tab:1},
showing the improvement from \modelname on LUKE by 2.6\% on TACRED, 1.0\% on TACRED-Revisit, 0.7\% on Re-TACRED, and 1.0\% on SemEval.
Overall, our \modelname method improves the effectiveness of RE significantly in terms of both F1-macro and F1-micro scores.
The above experimental results validate the effectiveness and generalization of our proposed method.

Among the baseline debiasing methods, Resample, Focal, CFIE cannot distill the entity bias in an entity-aware manner like ours.
Entity Mask leads to the loss of information, while our \modelname enables RE models to focus on the main effects of textual context without losing the entity information.
The superiority of \modelname highlights the importance of the causal inference based entity bias analysis for debiasing RE, which compares traditional likelihood-based predictions and hypothesized counterfactual ones to produce debiased predictions.
Besides, the proposed \modelname works in inference and thus can be employed on the previous already-trained models.
In this way, \modelname serves as a model-agnostic approach to enhance RE models without changing their training process.

\begin{table}[tb!]
	\centering
	\begin{adjustbox}{width=\linewidth}
		\begin{tabular}{@{}l c c | l c c @{}}
			\toprule
			\textbf{LUKE + \modelname }
			& \textbf{61.7}
			& \textbf{$\Delta$ }
			& \textbf{IRE + \modelname } & \textbf{64.4} & \textbf{$\Delta$}
			\\
			\midrule
			w/o \modelname & 58.8 & 2.9$\downarrow$ & w/o \modelname & 63.1 & 1.3$\downarrow$\\
			w/o EBM & 59.5 & 2.2$\downarrow$ &  w/o EBM & 63.4 & 1.0$\downarrow$ \\
			w/o LBM & 60.8 & 0.9$\downarrow$ & w/o LBM & 63.9 & 0.5$\downarrow$ \\
			w/o BSH & 60.1 & 1.6 $\downarrow$ & w/o BSH & 63.8 & 0.6$\downarrow$ \\
			\bottomrule
		\end{tabular}
	\end{adjustbox}

	\caption{Ablation study based on the TACRED dataset. The analyzed model components include entity bias mitigation operation (EBM), the label bias mitigation operation (LBM) and the beam search for hyper-parameters (BSH).
	`w/o' denotes `without'.
	$\downarrow$ denotes performance drop in terms of F1-macro scores.}\label{tab:abl}
	\vspace{-0.4cm}
\end{table}

\subsection{Analysis on Entity Bias}\label{sec:4-3}
Some work argues that RE models may rely on the entity mentions to make relation predictions instead of the textual context \cite{zhang2018graph,joshi-etal-2020-spanbert}.
The empirical results in \Cref{fig:5} validates this argument.
Regardless of whether the textual context exists or not, the baseline RE model makes the same predictions given only entity mentions on many instances.
The entity bias can mislead the RE models to make wrong predictions when the relation implied by the entity mentions does not exist in the textual context.

\begin{table*}[!ht]
	\centering
	\renewcommand\arraystretch{2.3}
	\begin{adjustbox}{width=\linewidth}

		\begin{tabular}{@{}l|c|c|c @{}}
			\toprule 
			\textbf{Input sentence}
			& \textbf{Original}
			& \textbf{Debiased}
			& \textbf{Counterfactual}
			\\ \midrule
			\multirow{2}{9cm}{More than 1,100 miles (1,770 kilometers) away, \underline{Alan Gross} passes his days in a \uwave{Cuban} military hospital, watching baseball on a small television or jamming with his jailers on a stringed instrument they gave him.} & 
            \multirow{2}{*}{origin \xmark}&
            \multirow{2}{*}{countries\_of\_residence \cmark}&
            
            \multirow{2}{*}{origin}

            \\
            &&
			\\
			\midrule
			
			\multirow{2}{9cm}{He said that according to his investigation, \underline{Bibi} drew the ire of fellow farmhands after a dispute in June 2009, when they refused to drink water she collected and she refused their demands that she convert to \uwave{Islam}.} & 
			
            \multirow{2}{*}{religion \xmark}&
            
            \multirow{2}{*}{no\_relation \cmark}&
            
            \multirow{2}{*}{religion}

            \\
            &&
			\\
			\midrule
			
			\multirow{2}{9cm}{\underline{ShopperTrak} also estimates foot traffic in the \uwave{U.S.} was 11.2 percent below what it would have been Sunday if the blizzard had not occurred and 13.9 percent below what it could have been Monday.}
			& 
			
            \multirow{2}{*}{country\_of\_headquarters \xmark}&
            
            \multirow{2}{*}{no\_relation \cmark}&
            
            \multirow{2}{*}{country\_of\_headquarters}

            \\
            &&
			\\
			\bottomrule
		\end{tabular}
	\end{adjustbox}
	\caption{A case study for IRE$_{\mathrm{RoBERTa}}$ and our \modelname on the relation extraction dataset TACRED.
	\underline{Underlines} and \uwave{wavy lines} highlight the subject and object entities respectively.
	We report the original prediction, the corresponding counterfactual prediction and the debiased prediction.}
	\label{tab:case}
	\vspace{-0.4cm}
\end{table*}

To evaluate whether RE models can generalize well to particularly challenging instances where relations implied by the entity mentions do not exist in the textual context, we propose a filtered evaluation setting, where we keep the test instances having the entity bias different from their ground-truth relations.
In this setting, RE models cannot overly rely on the entity mentions for RE, since the entity mentions no longer provide the superficial and spurious clues for the ground-truth relations.

We present the evaluation results on the filtered test set in \Cref{tab:chall}.
Our \modelname method consistently and substantially improves the effectiveness of LUKE and IRE on the filtered test set and outperforms the baseline methods by a significant margin, which validates the effectiveness and generalization of our method to mitigate the entity bias in the challenging cases.

\subsection{Evaluation on Fairness}
According to \citet{sweeney2019transparent}, the more imbalanced/skewed a prediction produced by a trained model is, the more unfair opportunities it gives over predefined categories, and the more unfairly discriminative the trained model is.
We thus follow previous work \cite{xiang2020learning,sweeney2019transparent,qian2021counterfactual} to use the metric – 
\textit{imbalance divergence} – to evaluate how imbalanced/skewed/unfair a prediction $P$ is :
\begin{equation}
    D(P, U) = JS(P \| U),
\end{equation}
where $D(\cdot)$ is defined as the distance between $P$ and the uniform distribution $U$.
Specifically, we use the \textit{JS} divergence as the distance metric since it is symmetric (i.e., $JS(P \| U) = JS(U \| P)$) and strictly scoped \cite{fuglede2004jensen}. 
Based on this, to evaluate the entity bias of a trained RE model, we average 
the following \textit{relative entity mention imbalance} (REI) measure over all the testing instances containing whichever entity mentions:
\begin{equation}
    \mathrm{REI} = \frac{1}{\mathcal{E}}\sum_{e \in \mathcal{E}}D(P(\{x | e \in x \land x \in \mathcal{D}\}), U),
\end{equation}
where $x$ is an input instance, $\mathcal{D}$ is the testing set, $P(x)$ is the prediction output, $e$ is an entity mention, and $\mathcal{E}$ is the corpus of entity mentions.
This metric captures the distance between all predictions and the fair uniform distribution $U$.

We follow the experimental settings in \Cref{sec:4-1} and report the fairness test in \Cref{tab:fair}.
The results show that our \modelname method reduces the imbalance metrics (lower is better) when employed on IRE$_{\mathrm{RoBERTa}}$ significantly and consistently, indicating that it is helpful to mitigate the entity bias.

\subsection{Ablation and Case Study}
We conduct ablation studies on \modelname to empirically examine the contribution of its main technical components. including the entity bias mitigation operation (EBM), the label bias mitigation operation (LBM) and the beam search for hyper-parameters (BSH).

We report the experimental results of the ablation study in \Cref{tab:abl}. 
We observe that removing our \modelname causes serious performance degradation.
This provides evidence that using our counterfactual framework for RE can explicitly mitigate biases to generalize better on unseen examples. 
Moreover, we observe that mitigating the two types of biases is consistently helpful for RE. 
The key reason is that the distilled label bias provides an instance-agnostic offset and the distilled entity bias provides an entity-aware one in the prediction space, which makes the RE models focus on extracting relations on the textual context without losing the entity information. 
Meanwhile, the beam search for hyper-parameters effectively finds two dynamic scaling factors
to amplify or shrink two biases, making the biases be mitigated properly and adaptively.

\Cref{tab:case} gives a qualitative comparison example
between \modelname and IRE$_{\mathrm{RoBERTa}}$ on TACRED. 
The results show that the state-of-the-art RE model IRE$_{\mathrm{RoBERTa}}$ returns the relations that do not exist in the textual context between the considered entities. 
For example, given ``\textit{\underline{Bibi} drew the ire of fellow farmhands after a dispute in June 2009, when they refused to drink water she collected and she refused their demands that she convert to \uwave{Islam}.}'', there is no relation between \textit{Bibi} and \textit{Islam} exists in the text but the baseline model believes that the relation between them is ``\textit{religion}''. 
The counterfactual prediction can account for this disappointing result, where given only the entity mentions \textit{Bibi} and \textit{Islam}, the RE model returns the relation ``\textit{religion}'' without any textual context. 
This implies that the model makes the prediction for the original input relying on the entity mentions, which leads to the wrong RE prediction.
Our \modelname method distills the biases through counterfactual predictions and mitigates the biases to distinguish the main effects from the textual context, which leads to the correct predictions as shown in \Cref{tab:case}.

Last but not least, we conduct experiments on the fairness of different models, and present respective results in the appendix.

\section{Conclusion}
We have designed a counterfactual analysis based method named \modelname to debias RE.
We distill the entity bias and mitigate the distilled biases with the help of our causal graph for RE, which is a road map for analyzing the RE models.
Based on the counterfactual analysis, we can analyze the side-effects of entity mentions in the RE and debias the models in an entity-aware manner.
Extensive experiments demonstrate that our methods can improve the effectiveness and generalization of RE.
Future work includes analyzing the effects of other factors that can cause bias in natural language processing.

\section*{Acknowledgement}
The authors would like to thank the anonymous reviewers for their discussion and feedback.

Muhao Chen and Wenxuan Zhou are supported by the National Science Foundation of United States Grant IIS 2105329, and by the DARPA MCS program under Contract No. N660011924033 with the United States Office Of Naval Research.
Except for Muhao Chen and Wenxuan Zhou, this paper is supported by NUS ODPRT Grant R252-000-A81-133 and Singapore Ministry of Education Academic Research Fund Tier 3 under MOEs official grant number MOE2017-T3-1-007.

\bibliography{acl2020}
\bibliographystyle{acl_natbib}

\end{document}